\pdfoutput=1

\documentclass[11pt]{article}

\usepackage[final]{acl}

\usepackage{times}
\usepackage{latexsym}

\usepackage[T1]{fontenc}

\usepackage[utf8]{inputenc}

\usepackage{microtype}

\usepackage{inconsolata}
\usepackage{orcidlink}

\newcommand{\FRorcid}{\orcidlink{0000-0002-1697-8586}}
\newcommand{\GSorcid}{\orcidlink{0009-0005-5221-0717}}

\title{Interlocking-free Selective Rationalization Through Genetic-based Learning}

\author{Federico Ruggeri\FRorcid \and
  Gaetano Signorelli\GSorcid \\
  DISI, University of Bologna \\
  \texttt{\{federico.ruggeri6, gaetano.signorelli2\}@unibo.it} \\
  }

\usepackage{graphicx}
\usepackage{booktabs}
\usepackage{amsmath}
\usepackage{cleveref}
\usepackage{xspace}
\usepackage{amsfonts}
\usepackage{algorithm,algorithmic}
\usepackage{multirow}

\newcommand{\gen}[0]{$g_\theta$\xspace}
\newcommand{\pred}[0]{$f_\omega$\xspace}
\newcommand{\method}[0]{GenSPP\xspace}

\begin{document}

\maketitle

\begin{abstract}

A popular end-to-end architecture for selective rationalization is the select-then-predict pipeline, comprising a generator to extract highlights fed to a predictor.
Such a cooperative system suffers from suboptimal equilibrium minima due to the dominance of one of the two modules, a phenomenon known as \textit{interlocking}.
While several contributions aimed at addressing interlocking, they only mitigate its effect, often by introducing feature-based heuristics, sampling, and ad-hoc regularizations.
We present \method, the first interlocking-free architecture for selective rationalization that does not require any learning overhead, as the above-mentioned. 
\method avoids interlocking by performing disjoint training of the generator and predictor via genetic global search.
Experiments on a synthetic and a real-world benchmark show that our model outperforms several state-of-the-art competitors.


\end{abstract}

\section{Introduction} \label{sec:intro}

\textit{Selective rationalization} is the process of learning by providing highlights (or rationales) as explanation, a type of explainable AI approach that has gained momentum in high-stakes scenarios~\cite{wiegreffe-marasovic-2021-teach-me}, such as fact-checking and legal analytics.
Highlights are a subset of input texts meant to be interpretable by a user and faithfully describe the inference process of a classification model~\cite{herrewijnen-etal-2024-rationale-survey}.
Among the several contributions, the select-then-predict (SPP) selective rationalization framework of~\citet{lei-etal-2016-rationalizing} has gained popularity due to its inherent property of defining a faithful self-explainable model.
In SPP, a classification model comprises a generator and a predictor.
The generator generates highlights from input texts, i.e., it selects a portion of input text tokens, which are fed to the predictor to address a task.
To define interpretable highlights, the generator performs discrete selections of input tokens while regularization objectives control the quality of generated highlights. 

This discretization process introduces an optimization issue between the generator and the predictor, hindering training stability and increasing the chances of falling into local minima, a phenomenon denoted as \textit{interlocking}~\cite{a2r}.
To account for this issue, several contributions have been proposed to facilitate information flow between the generator and predictor and avoid overfitting on sub-optimal highlights.
Notable examples include differentiable discretization via sampling~\cite{bao-etal-2018-deriving,bastings-etal-2019-interpretable}, weight sharing between generator and predictor~\cite{FR2022}, and external guidance via soft rationalization~\cite{a2r, huang-etal-2021-dmr,sha-etal-2023-infocal,g-rat-2024}.
However, these methods only mitigate interlocking by introducing ad-hoc regularization.
%

A few attempts have been proposed to eliminate interlocking.
These solutions either rely on feature-based heuristics to pre-train the generator~\cite{jain-etal-2020-learning} or partially address interlocking by introducing multiple independent training stages~\cite{3Stage}.
However, these methods present several limitations, including the use of heuristics for guiding the generator, limited information flow between the generator and the predictor, and introduce additional optimization issues.

We propose \textbf{Gen}etic-\textbf{SPP} (\method), the first selective rationalization framework that eliminates interlocking without requiring heuristics and architectural changes.
\method breaks interlocking by splitting the optimization process into two stages, optimized via genetic-based search.
First, a generator instance is defined independently of a given predictor.
Second, a predictor is trained from scratch while keeping the defined generator frozen.
By doing so, the generator instance is evaluated only via the training objective, without the need for additional regularizations to guide its learning to avoid interlocking.
Genetic-based search allows for local and global exploration of the generator's parameters, significantly reducing the risk of getting stuck into local minima.
Furthermore, genetic-based search does not require differentiable learning objectives, allowing for a more accurate model evaluation accounting for both classification performance and highlight quality.

We evaluate \method on two benchmarks: a controlled synthetic dataset that we introduce to assess selective rationalization frameworks and a popular real-world dataset on hate speech.
Experimental results show that \method achieves superior highlight quality while maintaining comparable classification performance.

To summarize, our contributions are:

\begin{itemize}
    \item We introduce \method, the first interlocking-free selective rationalization framework that does not require sub-optimal heuristics and additional regularizations.

    \item We design a robust evaluation objective to account for classification and rationalization capabilities equally.

    \item We build a novel controlled synthetic dataset to study selective rationalization frameworks.

    \item We carry out an extensive, robust, and reproducible experimental setting to compare \method with several competitive selective rationalization frameworks.
\end{itemize}

We make our data and code available for research.\footnote{
\url{https://github.com/nlp-unibo/gen-spp}
}

\section{Preliminaries} \label{sec:preliminaries}

We overview two fundamental concepts to understand our method: (i) selective rationalization and (ii) genetic-based search.

\subsection{Selective Rationalization}

Selective rationalization denotes a self-explainable classification model capable of extracting discrete highlights from an input text.
The typical architecture for selective rationalization is based on the select-then-predict (SPP) architecture~\cite{lei-etal-2016-rationalizing}.
In SPP, the classification model is split into a generator (\gen) and a predictor (\pred), where $\theta$ and $\omega$ are the parameter sets.
Given an input text $x = \{x^1, x^2, \dots, x^n \}$ comprising $n$ tokens and its corresponding ground-truth label $y$, the generator \gen produces a binary mask $m = g_\theta(x) = \{m^1, m^2, \dots, m^n \}$ where $m^i \in \{0, 1\}$.
The mask $m$ indicates which tokens of $x$ are selected.
We denote the mask generation process as \textit{rationalization}.
A masked input text $\tilde{x}$ is then defined by applying $m$ on $x$ as follows: $\tilde{x} = x \odot m$.
The masked text $\tilde{x}$ is fed to the predictor \pred for classification.
Generally, the selective rationalization architecture is trained to minimize the classification empirical error on an annotated dataset, without providing supervision on generated $m$.
This setting is often denoted as \textit{unsupervised rationalization}, which is formalized as follows:
\begin{equation} \label{eq:cooperative-rationalization}
    \mathcal{L}_t = \min_{\theta, \omega} \frac{1}{\vert \mathcal{D} \vert} \sum_{(x, y) \in \mathcal{D}} \mathcal{L}_{ce} \bigl (f_\omega(g_\theta(x) \odot x), y \bigr ), 
\end{equation}
where $\mathcal{D}$ is a textual dataset annotated for classification and $\mathcal{L}_{ce}$ is the classification loss.

\paragraph{Controlled Rationalization.}
A self-explainable model should produce meaningful highlights in addition to accurate predictions.
\citet{lei-etal-2016-rationalizing} introduced regularization objectives to prefer sparse and coherent highlights for better interpretability.
Formally, the regularizer is denoted as follows:
\begin{equation}
    \Omega(m) = \lambda_s \underbrace{\sum_{i=0}^n m^i}_{\mathcal{L}_{s}} + \lambda_c \underbrace{\sum_{i=1}^n \vert m^i - m^{i-1} \vert}_{\mathcal{L}_{c}},
\end{equation}
where $\mathcal{L}_{s}$ controls the level of sparsity (sparsity constraint), $\mathcal{L}_{c}$ reduces highlights fragmentation (contiguity constraint), and $\lambda_s, \lambda_c \in \mathbb{R}$ are scalar coefficients that balance the regularization.
Effectively, controlling the regularization effect of $\mathcal{L}_s$ to not outweigh $\mathcal{L}_t$ is non-trivial.
To simplify the optimization process, \citet{chang-etal-2020-invrat} relax the sparsity constraint to achieve a specific sparsity level: $\mathcal{L}_{s} = \vert \alpha - \frac{1}{n} \sum_{i=0}^n m^i \vert$, where $\alpha \in [0, 1]$ regulates the degree of sparsity.
By including the regularizer, Eq.~\ref{eq:cooperative-rationalization} can then be rewritten as follows:
%
\begin{equation} \label{eq:regularized-rationalization}
    \mathcal{L} = \mathcal{L}_t + \Omega(m)
\end{equation}
\paragraph{Interlocking.}

\citet{a2r} showed that when performing unsupervised rationalization in an end-to-end fashion, the selective rationalization architecture suffers from sub-optimal equilibrium minima.
This occurs when either the generator \gen or the predictor \pred are in a sub-optimal state.
If \gen is stuck on generating a sub-optimal $m$, \pred is fine-tuned on that $m$, further enforcing \gen to maintain that selection.
Similarly, if \pred is a remarkably bad predictor, it is further encouraged to exhibit lower classification error on a sub-optimal $m$ compared to the ground-truth one $m^*$.

\subsection{Genetic Algorithms}

Genetic Algorithms (GAs) constitute a class of search algorithms for finding optima in optimization problems.
They are based on population-based search relying on the concept of survival of the fittest~\cite{katoch-etal-2021-genetic-review}.
Formally, a population $\mathcal{P}$ contains a set of $I$ individuals, $\mathcal{P} = \{c_1, c_2, \dots, c_I \}$, where each individual $c \in \mathbb{R}^d$ is a parameter vector representing a candidate solution to the problem of interest.

Initially, a population $\mathcal{P}_0$ of $I$ individuals is initialized randomly to cover the solution search space.
The individuals are evaluated by a fitness function $h: \mathbb{R}^d \rightarrow \mathbb{R}$ that is the optimization objective of GAs.
A portion of individuals is then selected based on their fitness scores with selection probability $p^{sl}$.
An intermediate population $\tilde{\mathcal{P}}_0$ is built by generating individuals from selected ones, either by modifying a portion of individual parameters (\textit{mutation}) or by mixing parameters between individual pairs (\textit{crossover}).
We denote $p^m$ and $p^c$ the mutation and crossover probabilities, respectively.
The population for the next iteration $\mathcal{P}_i$ is built by performing a second individual selection phase, denoted as survival selection, to keep the number of individuals equal to $I$ across generations.
We denote $p^{su}$ the survival probability of each individual.
The population-based search is iterated for $G$ generations or stopped preemptively if a certain fitness score is reached.

\paragraph{Neuroevolution.}
GAs have been successfully applied to solve a wide variety of tasks~\cite{alhijawi-and-awajan-2024-genetic-survey}, including agent exploration \cite{conti-etal-2018-improving}, image processing, scheduling, clustering, natural language processing \cite{katoch-etal-2021-review}, training large neural networks \cite{Miikkulainen-etal-2019-evolving}, and, in particular, neural network optimization, known as \textit{neuroevolution}~\cite{galvan-and-mooney-neuroevolution-survey}.
Neuroevolution denotes the process of (i) neural network architecture search and (ii) parameter optimization by employing genetic algorithms.
In the second scenario, each individual $c$ in a population $\mathcal{P}$ denotes the parameters of a neural network.
In addition to having interesting properties, such as parallel computation and reduced likelihood of getting stuck into local minima, neuroevolution also shows correspondence with gradient descent, as proved by \citet{genetic-sgd-correspondence}.

\section{Related Work} \label{sec:related}
\citet{lei-etal-2016-rationalizing} introduce Rationalizing Neural Predictions (RNP), the first SPP framework, whereby the generator and predictor components are trained via reinforcement learning~\cite{williams-1992-reinforce}.
Several contributions have explored ways to improve RNP, including end-to-end optimizations, external guidance to mitigate spurious correlations, regularizations for faithful rationalization, and attempts to break interlocking.

\paragraph{Improved Optimization.}
\citet{bao-etal-2018-deriving} propose an end-to-end architecture by leveraging the Gumbel softmax trick~\cite{gumbel} for generating differentiable discrete masks $m$.
Similarly, \citet{bastings-etal-2019-interpretable} adopt rectified Kumaraswamy distributions to replace sampling from Bernoulli distributions. 
Parameterized sampling provides a regularization effect to mitigate interlocking, but it requires additional calibration effort to find the best trade-off between sampling stability and exploration.
In contrast, genetic-based search does not require sampling to define discrete selection masks and has superior optimization stability with respect to standard reinforcement learning algorithms~\cite{salimans-etal-2017-evolution-strategies}.
Contributions have also explored solutions to ease the learning process.
\citet{FR2022} propose to share embedding weights between the generator and predictor to increase information flow between the two modules.
\citet{dr-2023} employ different learning rates for \gen and \pred to mitigate selection mask overfitting.
\citet{MGR2023} use multiple generators to improve rationalization exploration to reduce the chance of interlocking.
While, in principle, some of these design choices, like weight sharing, may be included in our framework, they are not required as \method avoids interlocking.

\paragraph{External Guidance.}
Another class of contributions leverages information from the input text to guide selective rationalization.
\citet{a2r} define an attention-based predictor that performs soft selections to mitigate interlocking.
\citet{chang-etal-2019-car} propose a generator-discriminator adversarial training to learn class-wise highlights.
\citet{paranjape-etal-2020-information} propose a sparsity regularization objective based on information bottleneck to trade-off performance accuracy and highlight coherence.
\citet{huang-etal-2021-dmr} define a guider module that acts as a teacher for \pred and propose an embedding-based regularization between the embedded input $x$ and the generated highlight $\tilde{x}$ to guide \gen.
\citet{DARE2022} propose a mutual information regularization to exploit information from non-selected tokens by leveraging an additional predictor.
\citet{sha-etal-2023-infocal} introduce the InfoCal framework, where an additional predictor trained on the input text $x$ provides guidance through a regularization objective based on the information bottleneck principle.
\citet{dar-2023} use an additional predictor trained on the original texts and fixed during rationalization to guide \pred.
\citet{g-rat-2024} employ an end-to-end guidance module with information from the original input text to guide \pred while also providing importance scores for weighting tokens to guide \gen.
\citet{liu-etal-2024-mmi} propose an alternative to maximum mutual information, treating spurious features that correlate with class labels as noise.
In contrast to all these approaches, \method does not require the integration of additional neural modules and regularizations to guide \gen since genetic-based search alleviates selective rationalization from getting stuck into sub-optimal minima.

\paragraph{Breaking Interlocking.}
Few attempts have explored breaking interlocking.
\citet{jain-etal-2020-learning} employ importance score features derived from post-hoc explainable tools like LIME~\cite{ribeiro-etal-2016-lime} to first pre-train \gen.
Subsequently, \pred is trained on the dataset produced in the previous stage.
Compared to our work, the solution of \citet{jain-etal-2020-learning} has two limitations.
First, it requires external feature extraction tools that act as heuristics for training \gen in a supervised fashion.
Second, information learned when training \pred does not flow to \gen for improvement.
In contrast, the generator \gen in \method is trained via a heuristic fitness function that only involves learning objectives concerning classification performance and highlight quality ( Eq.~\ref{eq:regularized-rationalization}). 
A recent contribution is the 3-stage framework of \citet{3Stage} for multi-aspect rationalization~\cite{antognini-etal-2021-multidimensional, antonigni-and-faltings-2021}.
In the first stage, \gen and \pred are first trained end-to-end, and then \gen is discarded.
In the second stage, \pred is frozen, and a new generator is trained.
Likewise, in the third stage, the trained new generator is frozen while \pred is fine-tuned.
While this framework avoids interlocking by iteratively freezing \gen or \pred, it presents two main limitations.
First, it is not completely interlocking-free since interlocking may still occur in the first stage, leading to a sub-optimal \pred.
Second, it does not offer good guarantees for reaching an optimal solution due to two independent training stages.
In contrast, \method is interlocking-free, characterized by stable convergence properties due to global search. 

\section{Motivation} \label{sec:motivation}

We motivate our work by discussing how existing contributions only mitigate interlocking.
The analysis of~\cite {a2r} underlines that the quality of the selective rationalization solution strongly depends on the system's capability to avoid the interlocking effect, thus reducing the probability of incurring local minima during training.
Interlocking affects the following optimization problem:
\begin{equation} \label{eq:dual-minimization}
    \min_\theta \min_\omega \mathcal{L} (f_\omega(g_\theta(x) \odot x), y)
\end{equation}
A major cause of interlocking is the generation of a discrete binary mask $m$ to define a faithful and interpretable model.
The discretization of $m$ induces a discrepancy in how \gen and \pred learn during training.
As pointed out by~\citet{a2r}, \pred tends to overfit to a certain sub-optimal mask $m$, causing the interlocking.
More precisely, while the predictor's parameters $\omega$ change smoothly at each gradient step thanks to the continuous nature of the learning objective, the generator \gen contains a discrete function (i.e., rounding) that makes its policy a piecewise constant function with respect to its parameters $\theta$.
Even by applying smoothing techniques (e.g., sampling) to mitigate the issue and achieve differentiability, the generated binary mask $m$ might remain unchanged (or change too slowly) over multiple gradient steps, thus, leading \pred to overfit on $m$.

To address this issue, contributions have proposed sampling-based methods to allow for differentiable discretization~\cite{bao-etal-2018-deriving, bastings-etal-2019-interpretable}, external guidance by introducing an additional soft rationalization system~\cite{chang-etal-2019-car,a2r,sha-etal-2023-infocal,dar-2023,g-rat-2024}, multi-stage training procedures~\cite{MGR2023}, and weight sharing between \gen and \pred for increased information flow~\cite{FR2022}.
However, none of these methods solves interlocking, and the likelihood of rapidly falling into a local optimum is only mitigated at the cost of added optimization issues, such as increased variance.

Given the side effect caused by the unequal joint training of the two models via stochastic gradient descent (SGD), a logical and straightforward way to break the interlocking between \gen and \pred is to split the dual minimization problem of Eq.~\ref{eq:dual-minimization}.
Formally, let $\omega^*$ be the optimal predictor's parameters, and let $l$ be its optimal solution:
\begin{equation}
    l = \mathcal{L}(f_{\omega^*}(x), y)
\end{equation}
Eq.~\ref{eq:dual-minimization} can be reformulated as a disjoint training by minimizing:
\begin{equation} \label{eq:disjoint-2}
\begin{gathered}
    \min_\theta \Omega(m) \\
    s.t. \, \min_\omega \mathcal{L} \bigl(f_\omega(g_\theta(x) \odot x), y \bigr ) \leq l + \epsilon
\end{gathered}
\end{equation}
for a tolerance $\epsilon$.
This formulation is equivalent to finding the optimal highlight (according to the applied regularization), such that \pred achieves a comparable performance to a predictor trained on $x$, up to a certain level of approximation regulated by $\epsilon$.
Equivalently, \gen is trained to filter out uninformative information from input text $x$.
Given the structure of Eq.~\ref{eq:disjoint-2}, the disjoint optimization cannot be addressed via SGD and, therefore, we propose genetic algorithms to address the minimization problem.

\section{The GenSPP Framework} \label{sec:framework}

We introduce \method, a novel SPP framework optimized via genetic-based search.
\method presents several advantages over selective rationalization based on SGD.
First, \method is interlocking-free by splitting the optimization process into two stages (Eq.~\ref{eq:disjoint-2}): each individual $c$ embodies a different generator \gen, which is then evaluated through a unique predictor \pred.
Second, \method leverages genetic-based search, allowing for both local (via mutation) and global (via crossover) search in the $\theta$ parameter space to avoid local minima.
Third, genetic-based search does not require a differentiable learning objective, allowing for more accurate training regularizations.
We describe \method and discuss its advantages over other selective rationalization frameworks in detail.

\subsection{Method}
\method follows the same architecture of~\citet{lei-etal-2016-rationalizing} where hard rationalization is performed via rounding and is trained via neuroevolution.
In particular, individual evaluation is a two-stage process.
First, a population $\mathcal{P}$ of individuals, each representing a configuration of the generator's parameters, is defined.
Second, each individual is evaluated via a fitness function $h$.
In particular, a predictor is initialized from scratch for each individual $c$ and trained to minimize the task classification loss via SGD while keeping the parameters of $c$ frozen to avoid interlocking.
We compute $h$ on each trained model, and we build a new population by selecting individuals based on their fitness scores.
The process is iterated until convergence or a fixed budget of generations $G$ is reached.
\Cref{algo:genetic} summarizes \method algorithm.

\begin{algorithm}[!tb]
\caption{\method Algorithm}
\label{algo:genetic}
\begin{algorithmic}[1]
    \REQUIRE Population $\mathcal{P}$, fitness function $h$, selection probability $p^{sl}$, crossover probability $p^{c}$, mutation probability $p^m$, survival probability $p^{su}$, $G$ generations, task threshold $l$. \\
    \ENSURE Optimal individual $c^*$. \\
    \STATE Initialize $\mathcal{P}_0 = \{c_1, \dots, c_I \}$ of $I$ individuals
    \STATE Initialize memory weights $p_{\vert M \vert}$ = $p^0_{\vert M \vert}$
    \FOR{individual $c \in \mathcal{P}_0$}
        \STATE Train a predictor \pred to minimize $\mathcal{L}_t$
        \STATE Evaluate $c$ via fitness function $h$
    \ENDFOR
    \WHILE{current generation $g < G$}
        \STATE Determine crossover pairs with selection probability $p^c_i = \frac{h(c_i)}{\sum^I_j h(c_j)}$
        \STATE Generate $\frac{I}{2}$ new individuals via one-point crossover
        \STATE Perform mutation on newly generated individuals with mutation probability $p^m$
        \FOR{individual $c$ in generated individuals}
            \STATE Train a predictor \pred to minimize $\mathcal{L}_t$
            \STATE Evaluate $c$ via fitness function $h$
        \ENDFOR
        \STATE Perform survival selection to obtain $\mathcal{P}_{g+1}$
    \ENDWHILE 
\end{algorithmic}
\end{algorithm}

\subsection{Individual Evaluation} \label{sec:method:individual-evaluation}

\begin{figure}[!tb]
    \centering
    \includegraphics[clip, trim=1cm 10cm 1cm 10cm, width=1.0\columnwidth]{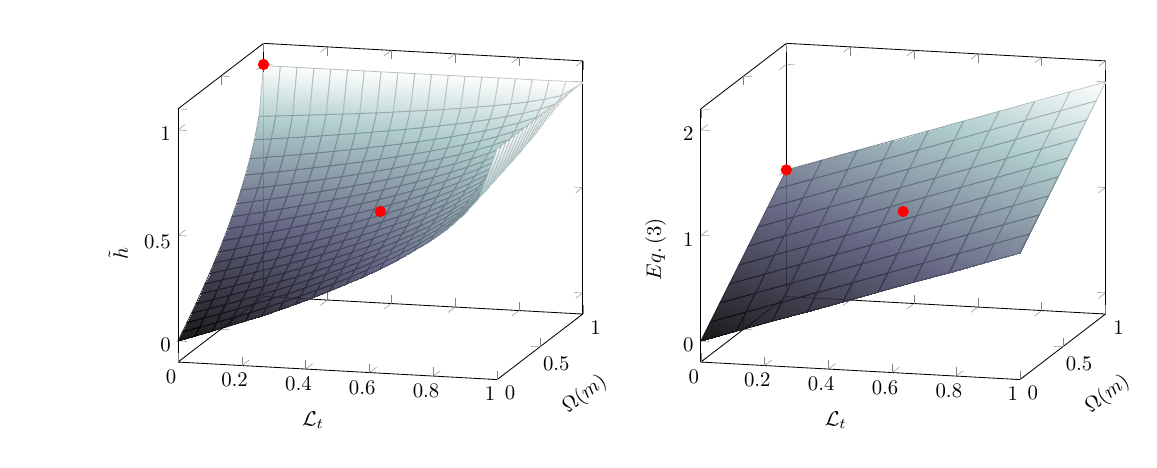}
    \caption{Loss landscape comparison between our fitness function $\tilde{h}$ (left) and the regularized selective rationalization objective (Eq.~\ref{eq:regularized-rationalization}). Red markers denote points (0.0, 1.0) and (0.5, 0.5), highlighting the difference between the two losses.}
    \label{fig:loss-comparison}
\end{figure}

We identify two major issues in Eq.~\ref{eq:regularized-rationalization} for model evaluation.
First, finding a balance between $\mathcal{L}_{t}$ and $\Omega(m)$ is non-trivial, potentially leading to sub-optimal solutions that only minimize one of the two.
Second, the joint learning formulation is not a reasonable candidate for optimization, collapsing substantially different solutions to the same cost value.
Consider two instances of the learning problem, one with $\mathcal{L}_{t} = 0.0$ and $\Omega(m) = 1.0$, and another with $\mathcal{L}_{t} = 0.5$ and $\Omega(m) = 0.5$.
Notably, both instances have the same average cost of $0.5$, but the first does not satisfy our objective of defining a faithful rationalization framework.
Fig.~\ref{fig:loss-comparison} compares the landscape of Eq.~\ref{eq:custom-fitness} to Eq.~\ref{eq:regularized-rationalization}.
Therefore, the two instances should be evaluated differently to favor solutions that are both accurate and interpretable.

To allow for more robust individual evaluation, we propose the following objective function:
\begin{equation}\label{eq:custom-fitness}
    \tilde{h} = 
    \begin{cases}
        1 - \mathcal{L},& \text{if } \mathcal{L}_{t} < l + \epsilon \\
        0,              & \text{otherwise}
    \end{cases}
\end{equation}
where $\mathcal{L} = \sqrt{(1 - \Omega(m)) \times (1 - min(\mathcal{L}_{t}, 1))}$.
To account for the maximization problem in genetic search, we define the fitness function $h$ in \method as follows:
\begin{equation} \label{eq:fitness-function}
    h = \frac{1}{\tilde{h} + \hat{\epsilon}},
\end{equation}
where $\hat{\epsilon}$ is a small constant to ensure computational stability.
Eq.~\ref{eq:custom-fitness} guides the learning process by initially favoring $\mathcal{L}_{t}$ and progressively shifting toward a state where $\mathcal{L}_{t}$ is stable while $\Omega(m)$ is optimized.
We do not require weight balancing since learning objectives are normalized and equally important.


\subsection{\method Genetic Algorithm}
We describe the genetic algorithm for training \method.
Given a population $\mathcal{P}_0$ of $I$ individuals, each representing a different generator instance, we perform individual selection and recombination as follows.
We initially evaluate $\mathcal{P}_0$ by computing the fitness score of each individual in the population.
We apply the roulette-wheel selection strategy, a stochastic process where individuals are sampled proportionally to their fitness score~\cite{lipowski-and-lipowska-2012-roulette-wheel}, to pair individuals for recombination.
In total, $\frac{I}{2}$ pairs are selected.
We employ one-point crossover~\cite{poli-etal-1998-onepoint-crossover} to generate $I$ new individuals from selected pairs.
This crossover strategy swaps parameters between two individuals by randomly choosing a swap point from a uniform distribution.
We then mutate each generated individual parameter with probability $p^m$ by inserting Gaussian noise.
The intermediate population $\tilde{\mathcal{P}_0}$ comprises the original $I$ individuals and the $I$ newly generated ones.
To build the population $\mathcal{P}_1$ of $I$ individuals for the next generation, we evaluate the fitness score of $\tilde{\mathcal{P}_0}$ and then perform survival selection via the half-elitism strategy~\cite{michalewicz-1996-genetic}.
In particular, we select the $\frac{I}{2}$ with the highest fitness score, while the remaining $\frac{I}{2}$ is sampled via roulette-wheel selection.

\subsection{Advantages}

Optimizing Eq.~\ref{eq:disjoint-2} via GAs introduces several advantages over selective rationalization based on SGD, which we discuss in detail.

\paragraph{Disjoint Training.}
A joint training of the selective rationalization system based on SGD involves a dependency between \gen and \pred: the quality of a highlight mask $m$ is also dependent on the quality of the current employed \pred (e.g., good masks may be evaluated badly if \pred has already overfitted to a previously generated mask).
In contrast, the proposed disjoint training allows the optimization of \gen by searching in the space of parameters that minimize $\Omega$, while yielding the highest performance in classification.
More precisely, the \pred depends on \gen, while the opposite does not hold.

\paragraph{Global Search.}
Population-based search in GAs reduces the chances of converging towards local minima, a common issue in optimization independently from interlocking.
Mutation and crossover offer two ways to perform local and global search space, respectively, alleviating the risk of getting stuck into a local optimum.

\paragraph{Non-differentiable Objective.}
Differentiable sampling (e.g., via Gumbel softmax~\cite{gumbel}) introduces noise, potentially making the optimization process of \gen unstable depending on the chosen sampling hyper-parameters.
In contrast, genetic-based search does not require gradient computation for optimization, ensuring a more robust training procedure.
Additionally, the optimization objective of \method (Eq.~\ref{eq:fitness-function}) can be designed without defining surrogate losses (Eq.~\ref{eq:custom-fitness}).
This is a crucial advantage of \method since it is not subject to dataset-specific hyperparameter-tuning (e.g., $\alpha$ in $\mathcal{L}_s$).
In contrast, SGD-based approaches require heavy fine-tuning to find a reasonable $\alpha$ value.

\section{Experimental Settings} \label{sec:experiments}

We compare \method to several competitors for unsupervised selective rationalization\footnote{We recall that ground-truth highlights are only used for model evaluation and not provided as input.} on two benchmarks.
We describe the data, models, and evaluation metrics in detail.
See Appendix~\ref{app:experimental-setting} for additional details.

\paragraph{Toy Dataset.}
We build and release a controlled toy dataset of random strings. 
We define three classification classes, each corresponding to a unique character-based highlight: \textit{aba}, \textit{baa}, \textit{abc}.
We design highlights to ensure that all their characters have to be selected in order to determine the corresponding class.
To avoid degenerate solutions in which only a portion of the highlight is sufficient for classification, we contaminate generated strings with randomly sampled chunks of other class highlights.
Lastly, we enforce that a single highlight is contained in each string.
Generated strings not compliant with the aforementioned rules are discarded.
We set the generated string length to 20 characters.
In total, we generate 10k random strings and split them into train (6.4k), validation (1.6k), and test (2k) partitions.

\paragraph{HateXplain Dataset.}
A dataset of $\sim$20k English posts from social media platforms like X and Gab~\cite{mathew-etal-2021-hatexplain}.
Each post is annotated from three different perspectives: hate speech (\textit{hate}, \textit{offensive}, \textit{normal}), the target community victim of hate speech, and the rationales which the labeling decision about hate speech is based on.
To account for annotation subjectivity, each post is annotated by at least three annotators~\cite{waseem-2016-racist, sap-etal-2022-annotators}.
We notice that annotations vary significantly among annotators regarding the number of selected tokens.
This might hinder rationalization evaluation since longer highlights might be preferred.
For this reason, we employ a majority voting strategy to merge annotators' highlights and identify top relevance tokens.
As a side effect, extracted ground-truth highlights are less cohesive.
We filter out texts longer than 30 tokens to reduce the computational overhead.
The dataset is split into train ($\sim$10k), validation ($\sim$1.3k), and test ($\sim$1.3k) partitions.
We consider hate speech as a binary classification problem by merging \textit{hate} and \textit{offensive} classes.

\paragraph{Models.}

We consider the architecture of~\citet{a2r} for all models, including ours, described as follows.
An input text $x$ is encoded via a frozen pre-trained embedding layer.
We use one-hot encoding for Toy and 25-dimension GloVe embeddings~\cite{pennington-etal-2014-glove} pre-trained on Twitter for HateXplain.
The generator \gen comprises a RNN layer with a dense layer on top for token selection.
The predictor \pred comprises a RNN layer followed by a max-pooling layer and a final linear layer for classification.
We set the RNN layer to a biGRU for baselines and GRU for \method, respectively.
We consider the following baselines.
FR~\cite{FR2022}, an end-to-end SPP framework using Gumbel softmax for discrete mask generation, where \gen and \pred share the same RNN layers.
MGR~\cite{MGR2023}, an SPP framework where multiple generators are considered to extract distinct highlights that are fed to a single predictor.
At inference time, only the first generator is considered since all generators eventually align on the same mask $m$.
MCD~\cite{mcd-2023}, a guidance-based SPP framework, where an additional predictor trained using the original input text $x$ is used to guide selective rationalization towards better highlights.
G-RAT~\cite{g-rat-2024}, a recent guidance-based SPP framework, where an attention-based soft SPP framework is used as guidance.

\paragraph{Evaluation Metrics.}
We focus on classification performance and rationalization quality~\cite{chang-etal-2019-car, a2r}.
Regarding classification performance, we report macro F1-score averaged over all classes (\textbf{Clf-F1}).
Regarding generated highlights, we report binary token-level F1-score (\textbf{Hl-F1}), selection ratio (\textbf{$R$}), and selection size (\textbf{$S$}).

\begin{table*}[!tb]
    \centering
    \small
    \resizebox{2\columnwidth}{!}{
    \begin{tabular}{l|rrrr|rrrr}
    \toprule
    & \multicolumn{4}{c}{\textbf{Toy}} & \multicolumn{4}{|c}{\textbf{HateXplain}} \\
     Model    & Clf-F1 $\uparrow$ & Hl-F1 $\uparrow$ & $R$ $\downarrow$ & $S$ $\downarrow$ & Clf-F1 $\uparrow$ & Hl-F1 $\uparrow$ & $R$ $\downarrow$ & $S$ $\downarrow$  \\ \midrule
     FR & $99.78_{\pm 0.20}$ & $54.07_{\pm 4.02}$ & $14.80_{\pm 0.24}$ & $2.96_{\pm 0.05}$ & $72.14_{\pm 1.12}$ & $31.15_{\pm 2.56}$ & $25.55_{\pm 0.72}$ & $3.46_{\pm 0.11}$ \\
     MGR & \boldmath $99.92_{\pm 0.04}$ & $50.34_{\pm 11.23}$ & $15.05_{\pm 0.80}$ & $3.01_{\pm 0.16}$ & $71.14_{\pm 1.16}$ & $29.38_{\pm 4.83}$ & $25.30_{\pm 1.04}$ & $3.42_{\pm 0.06}$ \\
     MCD & $99.90_{\pm 0.04}$ & $65.70_{\pm 3.76}$ & $15.18_{\pm 0.17}$ & $3.04_{\pm 0.03}$ & $70.37_{\pm 1.06}$ & $27.92_{\pm 1.66}$ & $25.07_{\pm 1.52}$ & $3.50_{\pm 0.20}$ \\
     G-RAT & $99.36_{\pm 0.82}$ & $50.22_{\pm 7.78}$ & $14.81_{\pm 0.51}$ & $2.96_{\pm 0.10}$ & \boldmath $73.85_{\pm 1.05}$ & $36.17_{\pm 1.62}$ & $24.68_{\pm 0.86}$ & $3.34_{\pm 0.08}$ \\
     \method (Ours) & $99.00_{\pm 0.25}$ & \boldmath $^{**}76.02_{\pm 0.64}$ & \boldmath $11.47_{\pm 0.49}$ & \boldmath $2.29_{\pm 0.08}$ & $69.71_{\pm 0.40}$ & \boldmath $^{**}42.62_{\pm 0.73}$ & \boldmath $6.51_{\pm 0.58}$ & \boldmath $0.75_{\pm 0.05}$ \\ \bottomrule
    \end{tabular}%
    }
    \caption{Benchmark evaluation test results. We report average macro F1-score (\textbf{Clf-F1}) for classification, while we report binary token-level F1-score (\textbf{Hl-F1}), selection rate (\textbf{R}) and size (\textbf{S}) for rationalization. Best results are highlighted in bold. \\
    \scriptsize{$(^{**}) \le 0.01$ denotes Wilcoxon statistical significance on the best baseline.}
    }
    \label{tab:unsupervised-rationalization:results}
\end{table*}

\begin{table*}[!tb]
    \centering
    \small
    \resizebox{2\columnwidth}{!}{
    \begin{tabular}{l|rrrr|rrrr}
    \toprule
    & \multicolumn{4}{c}{\textbf{Toy}} & \multicolumn{4}{|c}{\textbf{HateXplain}} \\
     Model    & Clf-F1 $\uparrow$ & Hl-F1 $\uparrow$ & $R$ $\downarrow$ & $S$ $\downarrow$ & Clf-F1 $\uparrow$ & Hl-F1 $\uparrow$ & $R$ $\downarrow$ & $S$ $\downarrow$  \\ \midrule
     FR & $99.85_{\pm 0.11}$ & $58.91_{\pm 3.18}$ & $14.57_{\pm 0.12}$ & $2.91_{\pm 0.02}$ & $71.00_{\pm 0.76}$ & $7.22_{\pm 2.29}$ & $26.85_{\pm 0.92}$ & $3.47_{\pm 0.08}$ \\
     MGR & $97.75_{\pm 4.25}$ & $37.45_{\pm 12.33}$ & $14.99_{\pm 0.42}$ & $3.00_{\pm 0.08}$ & $71.09_{\pm 1.00}$ & $14.45_{\pm 4.84}$ & $27.75_{\pm 1.36}$ & $3.57_{\pm 0.11}$ \\
     MCD & \boldmath $99.93_{\pm 0.06}$ & $62.94_{\pm 2.39}$ & $15.10_{\pm 0.66}$ & $3.02_{\pm 0.13}$ & $70.93_{\pm 0.95}$ & $13.88_{\pm 10.13}$ & $25.84_{\pm 1.87}$ & $3.46_{\pm 0.16}$ \\
     G-RAT & $99.85_{\pm 0.14}$ & $47.53_{\pm 12.77}$ & $14.56_{\pm 0.36}$ & $2.91_{\pm 0.07}$ & \boldmath $73.15_{\pm 0.35}$ & $34.33_{\pm 1.22}$ & $25.43_{\pm 0.87}$ & $3.40_{\pm 0.09}$ \\
     \method ($G = 100$) & $98.93_{\pm 0.47}$ & $70.52_{\pm 0.15}$ & $13.04_{\pm 0.12}$ & $2.60_{\pm 0.02}$ & $67.02_{\pm 0.57}$ & $39.89_{\pm 0.73}$ & $7.45_{\pm 0.48}$ & $0.96_{\pm 0.05}$ \\
     \method ($G = 150$) & $99.46_{\pm 0.36}$ & \boldmath $^{**}74.28_{\pm 0.61}$ & $10.11_{\pm 0.42}$ & $1.99_{\pm 0.04}$ & $69.89_{\pm 0.43}$ & \boldmath $^{**}42.81_{\pm 0.65}$ & \boldmath $6.74_{\pm 0.67}$ & \boldmath $0.87_{\pm 0.07}$ \\
     GenSPP$_{sk}$ ($G=100$) & $98.74_{\pm 0.43}$ & $63.45_{\pm 0.36}$ & \boldmath $8.03_{\pm 0.40}$ & \boldmath $1.58_{\pm 0.06}$ & $66.41_{\pm 0.35}$ & $35.52_{\pm 0.46}$ & $8.17_{\pm 0.62}$ & $1.06_{\pm 0.07}$ \\ \bottomrule
    \end{tabular}%
    }
    \caption{Synthetic skew test set results. We report average macro F1-score (\textbf{Clf-F1}) for classification, while we report binary token-level F1-score (\textbf{Hl-F1}) and selection rate (\textbf{R}) and size (\textbf{S}) for rationalization. Best results are highlighted in bold. \\
    \scriptsize{$(^{**}) \le 0.01$ denotes Wilcoxon statistical significance on the best baseline.}
    }
    \label{tab:skew:results}
\end{table*}

\section{Results} \label{sec:results}

We consider two sets of experiments.
The first evaluates models when trained from scratch to assess their capability to avoid local minima.
The second measures how good a method is at recovering from interlocking.
See Appendix~\ref{app:results} for additional results.

\paragraph{Benchmark Evaluation.}
Table~\ref{tab:unsupervised-rationalization:results} reports results.
We observe that \method significantly outperforms all competitors in selecting high-quality highlights (+10.3\% Hl-F1 in Toy and +6.5\% Hl-F1 in HateXplain), while reporting comparable classification performance.
Additionally, \method shows reduced variance across seed runs compared to competitors, especially in the Toy dataset, where MGR and G-RAT present notable instability.
Regarding highlight regularization, \method selects highlights that are more sparse and accurate compared to baseline models.
Interestingly, \method learns to not select any highlight for negative examples in HateXplain, while keeping valuable selections for positive examples, a flexibility that baseline models cannot achieve since they are subject to satisfy a certain sparsity threshold.
Overall, these results show the advantage of \method in performing a disjoint optimization problem via genetic-based search to break interlocking.

\paragraph{Synthetic Skewing.}
We follow~\citet{FR2022} and train a skewed \gen for $K = 10$ epochs using the classification label as supervision for selecting the first token $x^1$.
To evaluate \method on this experiment, we include one skewed individual in the initial population $\mathcal{P}_0$, while randomly initializing the remaining individuals.
We experiment with $G \in [100, 150]$ since convergence may require more time due to recombinations with the skewed individual in the earlier generations.
Additionally, to stress test \method, we consider a more degenerated setting where we initialize $\mathcal{P}_0$ with variants of the skew individual by adding Gaussian noise.
We denote this configuration as Gen-SPP$_{sk}$.
Table~\ref{tab:skew:results} reports results conducted on both datasets.
We observe that G-RAT and MCD are the best-performing baselines on HateXplain and Toy datasets, respectively.
In general, baseline models suffer from high variance, showing that these methods are not able to break the interlocking state in many seed runs.
In contrast, \method recovers from the degenerated state and outperforms baseline models, achieving comparable performance to the one reported in~\Cref{tab:unsupervised-rationalization:results}.
In particular, performing a parameter search with an increased budget (e.g., $G = 150$) leads to the best results.

\section{Discussion}

\paragraph{Computational Comparison.}
Breaking interlocking in \method comes with some limitations.
Intuitively, genetic-based search requires more computational time than solutions based on SGD since $I$ predictors are trained at each generation.
On average, a seed run of \method takes $\sim$36min in Toy and $\sim$78min in HateXplain.
In contrast, a seed run for baseline models requires $\sim$8min and $\sim$4min, respectively.
Nonetheless, we remark on two aspects regarding our implementation: (i) individuals are evaluated sequentially, and (ii) we make use of standard genetic operations for individual evaluation and selection.
More efficient implementations (e.g., allowing parallel computation of individuals) and advanced algorithms, such as the CMA-ES \cite{cma-es}, can significantly reduce convergence time.
We leave these improvements as future work.
This drawback is mitigated by two main properties of \method.
First, \method has low variance, avoiding, in principle, multiple seed runs for evaluation.
Second, global search via crossover allows for employing lighter and yet more efficient models.
Compared to competitors, \method has the same size as the smallest model (i.e., FR), which is 2-4x smaller than other baselines.

\paragraph{Genetic Search.}
Genetic search can be more time-consuming than other approaches like gradient-based methods. Since scalability issues are attributable to model complexity (number of parameters) rather than dataset size, they can be mitigated via parallel and efficient implementations \cite{cantupaz-and-goldberg-2000-efficient}. Model complexity is correlated with task complexity and is often addressed by increasing model parameterization. 
Genetic-based search can help reduce over-parameterization. 
Indeed, we shot that \method outperforms competitors despite having a smaller number of parameters, independently of the size of the dataset. 
Regarding text length (i.e., the number of tokens), token-level rationalization has O(n) time complexity (i.e., time scales linearly with tokens). 
However, we point out that selective rationalization for longer sequences is often accomplished at the sentence level to reduce task complexity \cite{antognini-etal-2021-multidimensional, hu-etal-2022-chef}.

\paragraph{Rationale Evaluation.}
In comparing models, sparsity and performance are equally important \cite{lei-etal-2016-rationalizing}. In particular, there is a preference for sparser models inspired by studies on human cognition \cite{hoefler-etal-2021-sparsity}.
The novelty of \method over gradient-based methods is that it does not require a specific sparsity threshold. In contrast, these methods present such a limitation to avoid unstable training regimes \cite{chang-etal-2020-invrat}. Like gradient-based methods, \method can be optimized for a certain sparsity threshold (see Tables \ref{tab:appendix:rationalization:toy} and \ref{tab:appendix:rationalization:hatexplain}).
Additionally, \method can be optimized for sparser solutions while maintaining high accuracy without additional constraints.






\section{Conclusions} \label{sec:conclusion}


We have introduced \method, the first selective rationalization framework that breaks interlocking via genetic-based search.
\method does not require differentiable surrogate learning objectives, additional regularization tuning, and architectural changes.
Our results on two benchmarks, a controlled synthetic one that we curate, and a real-world dataset for hate speech, show the advantage of \method, outperforming several competitors.
Furthermore, our robust evaluation underlines the increased variance that affects competitors' models, a phenomenon that was not sufficiently explored in selective rationalization.
Future research directions regard exploring more efficient genetic algorithms and implementations to reduce computational overhead and scale to more complex neural architectures.

\section*{Limitations}

\paragraph{Data.}
This study is based on two datasets, one of which is synthetic.
Some widely adopted datasets like Hotel Reviews \cite{wang-etal-2010-hotel-review} and Beer Reviews \cite{mcauley-etal-2021-beer} could be considered.
However, after a curated analysis, we excluded these datasets due to (i) data leakage between train, development and test splits; (ii) limited test set size (200-800 samples) which limits a robust and extensive evaluation of models; and (iii) single human annotations despite rationale selection being inherently subjective in these tasks.
Despite our efforts, we could not find other high-quality datasets for text classification with a relatively sufficient number of samples and multiple annotations for robust evaluation.
A broader analysis of \method on several datasets, including tasks other than text classification as in the ERASER benchmark \cite{deyoung-etal-2020-eraser}, could strengthen our contribution.

\paragraph{Models.}
All models follow a specific architecture, which is not the only possible one.
Our study could include other backbone architectures for a more exhaustive evaluation of selective rationalization frameworks.
However, while some existing contributions have explored more complex architectures like transformers \cite{g-rat-2024}, their effectiveness is still a matter of debate due to their sensitivity to interlocking \cite{MGR2023,liu-etal-2024-mmi}.

\paragraph{Algorithm Implementation.}
Our \method implementation does not leverage several optimizations available for genetic algorithms like parallel evaluation of individuals and distributed computing \cite{cantupaz-and-goldberg-2000-efficient}.
Consequently, we observed a notable computational overhead compared to gradient-based baselines.
More efficient implementations could be considered to reduce the overhead.

\paragraph{Task Complexity.}
The datasets used in this paper presented selective rationalization tasks where the number of tokens that could be selected is relatively small (up to 30 tokens).
More complex settings could be considered to further corroborate the advantages of genetic search for selective rationalization. 
However, we remark that the aim of our work is to propose a sound methodology to solve interlocking, and not to account for the limitations of genetic algorithms in large scale problems, which deserve a separate study.

\section*{Acknowledgments}
F. Ruggeri is partially supported by the project European Commission's NextGeneration EU programme, PNRR -- M4C2 -- Investimento 1.3, Partenariato Esteso, PE00000013 - ``FAIR - Future Artificial Intelligence Research'' -- Spoke 8 ``Pervasive AI’’ and by the European Union’s Justice Programme under Grant Agreement No. 101087342 for the project “Principles Of Law In National and European VAT”.

\bibliography{custom}

\appendix


\section{Experimental Settings} \label{app:experimental-setting}

\subsection{Data}

We report additional details regarding the presented datasets.

\paragraph{Toy Dataset.}
To assess the quality of our toy dataset, we evaluate string-matching baselines for selective rationalization. Intuitively, the baseline that selects the right highlight for each class should achieve perfect rationalization performance.
In contrast, other selections should lead to much lower selection performance.
We consider the following string-matching baselines: \{aba, baa, abc\}, \{abc, baa, aba\}, and \{ba, aa, bc\}.
The baselines achieve $100$\%, $33.33$\% and $53.57$\% Hl-F1 score, respectively.

\begin{figure}[!tb]
    \centering
    \includegraphics[clip, trim=6.5cm 10cm 6.2cm 10cm, width=0.5\textwidth]{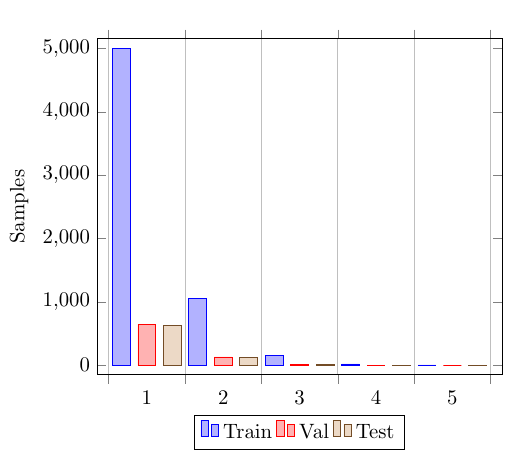}
    \caption{Number of contiguous highlights (i.e., connected token groups) in HateXplain.}
    \label{fig:data:hatexplain}
\end{figure}

\paragraph{HateXplain Dataset.}
Aggregating annotators' provided highlights via majority voting produces fragmented highlights.
Therefore, the contiguity constraint $\mathcal{L}_{c}$ may lead to sub-optimal solutions.
We compute the number of contiguous highlights in each example $x$ to systematically analyze the impact of our design choice (see Fig.~\ref{fig:data:hatexplain}).
Additionally, we compute the average highlight size and sparsity percentage.
On average, $S = 1.57_{\pm 2.52}$ which corresponds to $R = 0.12_{\pm 0.17}$.

\subsection{Training Setup}
We carry out a repeated train-and-test evaluation routine using the provided dataset partitions.
We evaluate models in five distinct seed runs.
We consider layer norm~\cite{ba-etal-2016-layernorm}, and early stopping on validation loss with patience set to 30 epochs as regularization methods.
We train models using batch size 64 and Adam optimizer~\cite{kingma-and-ba-2015-adam} with learning rate set to $10^{-3}$.
All baseline models are trained with SGD following Eq.~\ref{eq:regularized-rationalization} as training objective, where $\mathcal{L}_{ce}$ is the categorical cross-entropy.
We set $\lambda_s = 1.0$ and $\lambda_c = 2.0$ in the Toy dataset, while we set $\lambda_c = 0$ for HateXplain since highlights are inherently more fragmented (Fig.~\ref{fig:data:hatexplain}).
We set the sparsity threshold $\alpha = 0.15$ in the Toy dataset. This value of $\alpha$ encourages $\sum_{i=0}^{n} m^i = 3$, which is the length of all character-based highlights in the Toy dataset. Conversely, we set $\alpha = 0.22$ in HateXplain based on training data statistics of ground-truth highlights.

Regarding \method, we set $G = 100$ and $I = 50$, with mutation and crossover probabilities $p^m = p^c = 1.0$ and selection and survival rates $p^{sl} = p^{su} = 0.5$.
We perform mutation by adding a Gaussian noise sample from $\mathcal{N} (0.0, 0.05)$.
We train predictors during the genetic-based search for $3$ epochs with batch size $64$ and learning rate of $10^{-2}$.
We set evaluation tolerance $l + \epsilon$ equal to $0.1$ and $0.6$ for Toy and HateXplain case studies, respectively.

\subsection{Model Details}
\Cref{tab:appendix:hyperparameters} reports the full list of model hyper-parameters employed in our experiments, while \Cref{tab:appendix:hyperparameters:toy} and \Cref{tab:appendix:hyperparameters:hatexplain} report model configurations in Toy and HateXplain datasets, respectively.

\begin{table*}[!tb]
    \centering
    \begin{tabular}{l|l}
        \toprule
        Name & \multicolumn{1}{c}{Description} \\ \midrule
        \texttt{emb\_dim} & Input embedding dimension \\
        \texttt{emb\_type} & Pre-trained embedding matrix type \\
        \texttt{num\_classes} & Number of classification classes \\ \midrule
        \texttt{hidden\_size} & Number of units in RNN layers \\
        \texttt{cell\_type} & Type of RNN layer for encoding \\
        \texttt{num\_generators} & Number of generators in MGR \\ \midrule
        $\lambda_s$ & Coefficient for sparsity regularization $\mathcal{L}_s$ \\
        $\lambda_c$ & Coefficient for contiguity regularization $\mathcal{L}_c$ \\
        $\lambda_{kl}$ & Kullback-Lieber divergence coefficient in MCD \\
        $\lambda_{jsd}$ & Jensen-Shannon divergence coefficient in G-RAT \\
        $\lambda_{g}$ & Guider coefficient in G-RAT \\
        \texttt{pretrain} & Number of guider pre-training epochs in G-RAT \\
        $g\_decay$ & Guider regularization decay coefficient in G-RAT \\
        $\sigma$ & Attention noise in guider model in G-RAT \\
        $G$ & Number of genetic-based search generations in \method \\
        $I$ & Population size in \method \\
        $p^m$ & Mutation probability in \method \\
        $p^c$ & Crossover probability in \method \\
        $p^{sl}$ & Selection probability in \method \\
        $p^{su}$ & Survival probability in \method \\ \bottomrule
    \end{tabular}
    \caption{List of hyper-parameters in employed selective rationalize models.}
    \label{tab:appendix:hyperparameters}
\end{table*}

\begin{table*}[!tb]
    \centering
    \begin{tabular}{l|p{3.2cm}|p{3.5cm}|p{3.2cm}|p{3.1cm}}
        \toprule
        Model & \multicolumn{1}{c|}{General} & \multicolumn{1}{c|}{\gen} & \multicolumn{1}{c|}{\pred} & \multicolumn{1}{c}{Learning} \\ \midrule
        FR  & \begin{tabular}{l} \texttt{emb\_dim}: $25$\\ \texttt{emb\_type}: 1-hot\\ \texttt{num\_classes}: $3$ \end{tabular} & \begin{tabular}{l}
         \texttt{hidden\_size}: $8$ \\ \texttt{cell\_type}: biGRU \end{tabular} & \begin{tabular}{l} \texttt{hidden\_size}: $8$\\ \texttt{cell\_type}: biGRU \end{tabular} & \begin{tabular}{l} $\lambda_s$: $1.0$\\ $\lambda_c$: $1.0$ \end{tabular} \\ \midrule
        MGR  & \begin{tabular}{l} \texttt{emb\_dim}: $25$ \\ \texttt{emb\_type}: 1-hot \\ \texttt{num\_classes}: $3$ \end{tabular} & \begin{tabular}{l}
         \texttt{hidden\_size}: $8$ \\ \texttt{cell}: biGRU \\ \texttt{num\_generators}: $3$ \end{tabular} & \begin{tabular}{l} \texttt{hidden\_size}: $8$\\ \texttt{cell}: biGRU \end{tabular} & \begin{tabular}{l} $\lambda_s$: $1.0$\\ $\lambda_c$: $1.0$ \end{tabular} \\ \midrule
         MCD  & \begin{tabular}{l} \texttt{emb\_dim}: $25$\\ \texttt{emb\_type}: 1-hot\\ \texttt{num\_classes}: $3$ \end{tabular} & \begin{tabular}{l}
         \texttt{hidden\_size}: $8$ \\ \texttt{cell\_type}: biGRU \end{tabular} & \begin{tabular}{l} \texttt{hidden\_size}: $8$\\ \texttt{cell\_type}: biGRU \end{tabular} & \begin{tabular}{l} $\lambda_s$: $1.0$\\ $\lambda_c$: $1.0$ \\ $\lambda_{kl}$: $1.0$ \end{tabular} \\ \midrule
         G-RAT  & \begin{tabular}{l} \texttt{emb\_dim}: $25$\\ \texttt{emb\_type}: 1-hot\\ \texttt{num\_classes}: $3$ \end{tabular} & \begin{tabular}{l}
         \texttt{hidden\_size}: $8$ \\ \texttt{cell\_type}: biGRU \end{tabular} & \begin{tabular}{l} \texttt{hidden\_size}: $8$\\ \texttt{cell\_type}: biGRU \end{tabular} & \begin{tabular}{l} $\lambda_s$: $1.0$\\ $\lambda_c$: $1.0$ \\ $\lambda_{jsd}$: $1.0$ \\ $\lambda_{g}$: $1.0$ \\ \texttt{pretrain}: $10$ \\ \texttt{g\_decay}: $1e^{-05}$ \\ $\sigma$: $1.0$ \end{tabular} \\ \midrule
         \method & \begin{tabular}{l} \texttt{emb\_dim}: $25$\\ \texttt{emb\_type}: 1-hot\\ \texttt{num\_classes}: $3$ \end{tabular} & \begin{tabular}{l}
         \texttt{hidden\_size}: $8$ \\ \texttt{cell\_type}: GRU \end{tabular} & \begin{tabular}{l} \texttt{hidden\_size}: $8$\\ \texttt{cell}: GRU \end{tabular} & \begin{tabular}{l} $G$: $100$\\ $I$: $50$ \\ $p^m$: 1.0 \\ $p^c$: 1.0 \\ $p^{sl}$: 0.5 \\ $p^{su}$: 0.5 \end{tabular} \\ \bottomrule
    \end{tabular}
    \caption{Model hyper-parameters for Toy dataset.}
    \label{tab:appendix:hyperparameters:toy}
\end{table*}

\begin{table*}[!tb]
    \centering
    \begin{tabular}{l|p{3.2cm}|p{3.5cm}|p{3.2cm}|p{3.1cm}}
        \toprule
        Model & \multicolumn{1}{c|}{General} & \multicolumn{1}{c|}{\gen} & \multicolumn{1}{c|}{\pred} & \multicolumn{1}{c}{Learning} \\ \midrule
        FR  & \begin{tabular}{l} \texttt{emb\_dim}: $25$\\ \texttt{emb\_type}: GloVe\\ \texttt{num\_classes}: $2$ \end{tabular} & \begin{tabular}{l}
         \texttt{hidden\_size}: $16$ \\ \texttt{cell\_type}: biGRU \end{tabular} & \begin{tabular}{l} \texttt{hidden\_size}: $16$\\ \texttt{cell}: biGRU \end{tabular} & \begin{tabular}{l} $\lambda_s$: $1.0$\\ $\lambda_c$: $0.0$ \end{tabular} \\ \midrule
        MGR  & \begin{tabular}{l} \texttt{emb\_dim}: $25$ \\ \texttt{emb\_type}: GloVe \\ \texttt{num\_classes}: $2$ \end{tabular} & \begin{tabular}{l}
         \texttt{hidden\_size}: $16$ \\ \texttt{cell\_type}: biGRU \\ \texttt{num\_generators}: $3$ \end{tabular} & \begin{tabular}{l} \texttt{hidden\_size}: $16$\\ \texttt{cell}: biGRU \end{tabular} & \begin{tabular}{l} $\lambda_s$: $1.0$\\ $\lambda_c$: $0.0$ \end{tabular} \\ \midrule
         MCD  & \begin{tabular}{l} \texttt{emb\_dim}: $25$\\ \texttt{emb\_type}: GloVe\\ \texttt{num\_classes}: $2$ \end{tabular} & \begin{tabular}{l}
         \texttt{hidden\_size}: $16$ \\ \texttt{cell\_type}: biGRU \end{tabular} & \begin{tabular}{l} \texttt{hidden\_size}: $16$\\ \texttt{cell\_type}: biGRU \end{tabular} & \begin{tabular}{l} $\lambda_s$: $1.0$\\ $\lambda_c$: $0.0$ \\ $\lambda_{kl}$: $1.0$ \end{tabular} \\ \midrule
         G-RAT  & \begin{tabular}{l} \texttt{emb\_dim}: $25$\\ \texttt{emb\_type}: GloVe\\ \texttt{num\_classes}: $2$ \end{tabular} & \begin{tabular}{l}
         \texttt{hidden\_size}: $16$ \\ \texttt{cell\_type}: biGRU \end{tabular} & \begin{tabular}{l} \texttt{hidden\_size}: $16$\\ \texttt{cell\_type}: biGRU \end{tabular} & \begin{tabular}{l} $\lambda_s$: $1.0$\\ $\lambda_c$: $0.0$ \\ $\lambda_{jsd}$: $1.0$ \\ $\lambda_{g}$: $2.5$ \\ \texttt{pretrain}: $10$ \\ \texttt{g\_decay}: $1e^{-05}$ \\ $\sigma$: $1.0$ \end{tabular} \\ \midrule
         \method & \begin{tabular}{l} \texttt{emb\_dim}: $25$\\ \texttt{emb\_type}: GloVe\\ \texttt{num\_classes}: $2$ \end{tabular} & \begin{tabular}{l}
         \texttt{hidden\_size}: $16$ \\ \texttt{cell\_type}: GRU \end{tabular} & \begin{tabular}{l} \texttt{hidden\_size}: $16$\\ \texttt{cell}: GRU \end{tabular} & \begin{tabular}{l} $G$: $100$\\ $I$: $50$ \\ $p^m$: 1.0 \\ $p^c$: 1.0 \\ $p^{sl}$: 0.5 \\ $p^{su}$: 0.5 \end{tabular} \\ \bottomrule
    \end{tabular}
    \caption{Model hyper-parameters for HateXplain dataset.}
    \label{tab:appendix:hyperparameters:hatexplain}
\end{table*}

\subsection{Hardware and Implementation Details}
For our experiments, we implemented all baselines and methods in PyTorch~\cite{paszke-etal-2019-pytorch}, relying on open-source frameworks like PyTorch Lightning \cite{falcon-etal-2019-pytorch-lightning}.
We will release all of the code and data to reproduce our experiments in an MIT-licensed public repository.
All experiments were run on a private machine with an NVIDIA 3060Ti GPU with 8 GB dedicated VRAM.

\section{Results} \label{app:results}

We report additional experimental results for each presented experiment.

\paragraph{Benchmark Evaluation.}
\Cref{tab:appendix:rationalization:toy} and \Cref{tab:appendix:rationalization:hatexplain} report extensive results conducted on Toy and HateXplain datasets, respectively.
In addition to baseline models, we consider a random baseline to assess the complexity of the rationalization task.

\begin{table*}[!tb]
    \centering
    \small
    \begin{tabular}{lrrrr}
    \toprule
        Model    & Clf-F1 $\uparrow$ & Hl-F1 $\uparrow$ & $R$ $\downarrow$ & $S$ $\downarrow$ \\ \midrule
        FR ($\alpha = 0.10$) & $99.14_{\pm 1.23}$ & $50.23_{\pm 8.32}$ & $9.74_{\pm 0.47}$ & $1.95_{\pm 0.09}$ \\
        MGR ($\alpha = 0.10$) & $99.56_{\pm 0.21}$ & $40.02_{\pm 8.90}$ & $10.03_{\pm 0.40}$ & $2.01_{\pm 0.08}$ \\
        MCD ($\alpha = 0.10$) & $99.89_{\pm 0.05}$ & $62.89_{\pm 2.20}$ & $10.12_{\pm 0.38}$ & $2.02_{\pm 0.08}$ \\
        G-RAT ($\alpha = 0.10$) & $99.42_{\pm 0.94}$ & $50.33_{\pm 10.34}$ & $9.98_{\pm 0.21}$ & $2.00_{\pm 0.04}$ \\ \midrule
        FR ($\alpha = 0.15$) & $99.78_{\pm 0.20}$ & $54.07_{\pm 4.02}$ & $14.80_{\pm 0.24}$ & $2.96_{\pm 0.05}$ \\
        MGR ($\alpha = 0.15$) & $99.92_{\pm 0.04}$ & $50.34_{\pm 11.23}$ & $15.05_{\pm 0.80}$ & $3.01_{\pm 0.16}$ \\
        MCD ($\alpha = 0.15$) & $99.90_{\pm 0.04}$ & $65.70_{\pm 3.76}$ & $15.18_{\pm 0.17}$ & $3.04_{\pm 0.03}$ \\
        G-RAT ($\alpha = 0.15$) & $99.36_{\pm 0.82}$ & $50.22_{\pm 7.78}$ & $14.81_{\pm 0.51}$ & $2.96_{\pm 0.10}$ \\  \bottomrule
    \end{tabular}
    \caption{Test results on Toy when varying sparsity threshold $\alpha$. 
    }
    \label{tab:appendix:rationalization:toy}
\end{table*}

\begin{table*}[!tb]
    \centering
    \small
    \begin{tabular}{lrrrr}
    \toprule
        Model    & Clf-F1 $\uparrow$ & Hl-F1 $\uparrow$ & $R$ $\downarrow$ & $S$ $\downarrow$ \\ \midrule
        FR ($\alpha = 0.10$) & $70.80_{\pm 1.15}$ & $22.52_{\pm 14.64}$ & $13.02_{\pm 0.87}$ & $1.57_{\pm 0.07}$ \\
        MGR ($\alpha = 0.10$) & $69.74_{\pm 1.81}$ & $28.13_{\pm 10.99}$ & $13.76_{\pm 0.58}$ & $1.64_{\pm 0.06}$ \\
        MCD ($\alpha = 0.10$) & $68.52_{\pm 2.79}$ & $21.17_{\pm 15.03}$ & $12.96_{\pm 0.57}$ & $1.65_{\pm 0.03}$ \\
        G-RAT ($\alpha = 0.10$) & $71.33_{\pm 1.14}$ & $40.40_{\pm 3.25}$ & $13.00_{\pm 0.50}$ & $1.58_{\pm 0.07}$ \\ \midrule
        FR ($\alpha = 0.16$) & $71.90_{\pm 1.47}$ & $27.34_{\pm 13.41}$ & $19.31_{\pm 1.35}$ & $2.49_{\pm 0.12}$ \\
        MGR ($\alpha = 0.16$) & $71.03_{\pm 0.70}$ & $31.57_{\pm 5.60}$ & $20.28_{\pm 1.25}$ & $2.54_{\pm 0.08}$ \\
        MCD ($\alpha = 0.16$) & $70.03_{\pm 0.97}$ & $25.60_{\pm 6.98}$ & $19.65_{\pm 0.87}$ & $2.60_{\pm 0.07}$ \\
        G-RAT ($\alpha = 0.16$) & $71.68_{\pm 1.23}$ & $38.45_{\pm 2.31}$ & $19.73_{\pm 0.78}$ & $2.52_{\pm 0.04}$ \\ \midrule
        FR ($\alpha = 0.22$) & $72.14_{\pm 1.12}$ & $31.15_{\pm 2.56}$ & $25.55_{\pm 0.72}$ & $3.46_{\pm 0.11}$ \\
        MGR ($\alpha = 0.22$) & $71.14_{\pm 1.16}$ & $29.38_{\pm 4.83}$ & $25.30_{\pm 1.04}$ & $3.42_{\pm 0.06}$ \\
        MCD ($\alpha = 0.22$) & $70.37_{\pm 1.06}$ & $27.92_{\pm 1.66}$ & $25.07_{\pm 1.52}$ & $3.50_{\pm 0.20}$ \\
        G-RAT ($\alpha = 0.22$) & $73.85_{\pm 1.05}$ & $36.17_{\pm 1.62}$ & $24.68_{\pm 0.86}$ & $3.34_{\pm 0.08}$ \\ \midrule
        FR ($\alpha = 0.28$) & $73.09_{\pm 0.75}$ & $29.41_{\pm 1.32}$ & $31.08_{\pm 1.75}$ & $4.35_{\pm 0.14}$ \\
        MGR ($\alpha = 0.28$) & $72.41_{\pm 0.95}$ & $27.03_{\pm 3.28}$ & $31.11_{\pm 1.72}$ & $4.33_{\pm 0.12}$ \\
        MCD ($\alpha = 0.28$) & $70.26_{\pm 1.15}$ & $25.98_{\pm 0.76}$ & $30.29_{\pm 0.71}$ & $4.34_{\pm 0.14}$ \\
        G-RAT ($\alpha = 0.28$) & $73.60_{\pm 0.71}$ & $32.20_{\pm 0.96}$ & $31.26_{\pm 0.80}$ & $4.36_{\pm 0.11}$ \\ \bottomrule
    \end{tabular}
    \caption{Test results on HateXplain when varying sparsity threshold $\alpha$.
    }
    \label{tab:appendix:rationalization:hatexplain}
\end{table*}

\paragraph{Synthetic Skewing.}
\Cref{tab:appendix:skew} reports synthetic skew results when considering $K \in [5, 10, 15, 20]$.

\begin{table*}[!tb]
    \centering
    \small
    \begin{tabular}{l|l|rrrr|rrrr}
    \toprule
    & & \multicolumn{4}{c}{\textbf{Toy}} & \multicolumn{4}{|c}{\textbf{HateXplain}} \\
     & Model    & Clf-F1 $\uparrow$ & Hl-F1 $\uparrow$ & $R$ $\downarrow$ & $S$ $\downarrow$ & Clf-F1 $\uparrow$ & Hl-F1 $\uparrow$ & $R$ $\downarrow$ & $S$ $\downarrow$  \\ \midrule
     \multirow{4}{*}{\rotatebox[origin=c]{90}{$K = 5$}} & FR & $99.94_{\pm 0.06}$ & $50.32_{\pm 9.81}$ & $14.62_{\pm 0.32}$ & $2.92_{\pm 0.06}$ & $70.86_{\pm 1.45}$ & $14.91_{\pm 7.50}$ & $27.35_{\pm 1.01}$ & $3.48_{\pm 0.08}$ \\
     & MGR & $99.08_{\pm 1.54}$ & $41.26_{\pm 18.58}$ & $14.96_{\pm 0.68}$ & $2.99_{\pm 0.14}$ & $72.85_{\pm 0.77}$ & $21.84_{\pm 9.56}$ & $27.62_{\pm 1.13}$ & $3.42_{\pm 0.12}$ \\
     & MCD & $99.94_{\pm 0.04}$ & $65.18_{\pm 5.43}$ & $15.12_{\pm 0.45}$ & $3.02_{\pm 0.09}$ & $70.02_{\pm 1.56}$ & $22.72_{\pm 9.03}$ & $25.54_{\pm 1.32}$ & $3.53_{\pm 0.09}$ \\
     & G-RAT & $99.20_{\pm 1.37}$ & $46.79_{\pm 12.37}$ & $14.91_{\pm 0.13}$ & $2.98_{\pm 0.03}$ & $73.34_{\pm 0.34}$ & $33.51_{\pm 1.20}$ & $26.31_{\pm 0.99}$ & $3.45_{\pm 0.11}$ \\ \midrule
     \multirow{4}{*}{\rotatebox[origin=c]{90}{$K = 10$}} & FR & $99.85_{\pm 0.11}$ & $58.91_{\pm 3.18}$ & $14.57_{\pm 0.12}$ & $2.91_{\pm 0.02}$ & $71.00_{\pm 0.76}$ & $7.22_{\pm 2.29}$ & $26.85_{\pm 0.92}$ & $3.47_{\pm 0.08}$ \\
     & MGR & $97.75_{\pm 4.25}$ & $37.45_{\pm 12.33}$ & $14.99_{\pm 0.42}$ & $3.00_{\pm 0.08}$ & $71.09_{\pm 1.00}$ & $14.45_{\pm 4.84}$ & $27.75_{\pm 1.36}$ & $3.57_{\pm 0.11}$ \\
     & MCD & $99.93_{\pm 0.06}$ & $62.94_{\pm 2.39}$ & $15.10_{\pm 0.66}$ & $3.02_{\pm 0.13}$ & $70.93_{\pm 0.95}$ & $13.88_{\pm 10.13}$ & $25.84_{\pm 1.87}$ & $3.46_{\pm 0.16}$ \\
     & G-RAT & $99.85_{\pm 0.14}$ & $47.53_{\pm 12.77}$ & $14.56_{\pm 0.36}$ & $2.91_{\pm 0.07}$ & $73.15_{\pm 0.35}$ & $34.33_{\pm 1.22}$ & $25.43_{\pm 0.87}$ & $3.40_{\pm 0.09}$ \\ \midrule
    \multirow{4}{*}{\rotatebox[origin=c]{90}{$K = 15$}} & FR & $99.85_{\pm 0.15}$ & $51.44_{\pm 10.06}$ & $14.90_{\pm 0.59}$ & $2.98_{\pm 0.12}$ & $70.95_{\pm 1.51}$ & $8.21_{\pm 1.49}$ & $25.85_{\pm 1.54}$ & $3.39_{\pm 0.13}$ \\
     & MGR & $93.29_{\pm 11.14}$ & $22.10_{\pm 8.79}$ & $15.04_{\pm 0.48}$ & $3.01_{\pm 0.10}$ & $72.35_{\pm 0.69}$ & $10.90_{\pm 5.51}$ & $25.99_{\pm 1.66}$ & $3.42_{\pm 0.07}$ \\
     & MCD & $99.91_{\pm 0.07}$ & $62.78_{\pm 2.01}$ & $14.94_{\pm 0.33}$ & $2.99_{\pm 0.07}$ & $69.58_{\pm 0.81}$ & $11.04_{\pm 8.03}$ & $25.37_{\pm 1.40}$ & $3.41_{\pm 0.06}$ \\
     & G-RAT & $99.84_{\pm 0.21}$ & $42.84_{\pm 8.17}$ & $14.75_{\pm 0.47}$ & $2.95_{\pm 0.09}$ & $73.55_{\pm 0.32}$ & $33.43_{\pm 2.25}$ & $26.12_{\pm 1.80}$ & $3.36_{\pm 0.13}$ \\ \midrule
     \multirow{4}{*}{\rotatebox[origin=c]{90}{$K = 20$}} & FR & $99.63_{\pm 0.45}$ & $48.51_{\pm 13.52}$ & $14.49_{\pm 0.36}$ & $2.90_{\pm 0.07}$ & $71.30_{\pm 1.25}$ & $8.53_{\pm 2.21}$ & $27.18_{\pm 0.87}$ & $3.54_{\pm 0.08}$ \\
     & MGR & $89.05_{\pm 11.15}$ & $18.53_{\pm 5.19}$ & $16.07_{\pm 0.64}$ & $3.21_{\pm 0.13}$ & $71.30_{\pm 1.11}$ & $13.26_{\pm 4.31}$ & $27.31_{\pm 0.42}$ & $3.53_{\pm 0.11}$ \\
     & MCD & $99.91_{\pm 0.08}$ & $62.26_{\pm 2.50}$ & $14.80_{\pm 0.35}$ & $2.96_{\pm 0.07}$ & $69.75_{\pm 1.35}$ & $16.58_{\pm 9.54}$ & $26.44_{\pm 2.40}$ & $3.48_{\pm 0.20}$ \\
     & G-RAT & $66.95_{\pm 40.28}$ & $29.43_{\pm 10.21}$ & $21.43_{\pm 8.72}$ & $4.29_{\pm 1.74}$ & $73.57_{\pm 0.55}$ & $33.71_{\pm 1.06}$ & $26.53_{\pm 0.88}$ & $3.47_{\pm 0.07}$ \\ \bottomrule
    \end{tabular}
    \caption{Synthetic skew experiment results when varying skew pre-training epochs $K$. 
    }
    \label{tab:appendix:skew}
\end{table*}

\paragraph{Running Time and Model Size.}
\Cref{tab:appendix:time-and-size} reports training running time and model size for each selective rationalization evaluated in our experiments.
It is worth noting that for \method, we only report \pred trainable parameters, which are the only ones trained during individual evaluation.
If we consider \gen parameters, the \method size equals the one of FR.



\begin{table*}[!tb]
    \centering
    \small
    \begin{tabular}{lrrrr}
        \toprule
        Model & Single (min.) & Total (min.) & No. Parameters & Memory (MB) \\ \midrule
        \multicolumn{5}{c}{Toy} \\ \midrule
        FR & $7.41_{\pm 2.68}$ & $38.42$ & 1797 & 0.010 \\
        MGR & $8.94_{\pm 2.67}$ & $46.04$ & 7001 & 0.039 \\ 
        MCD & $6.55_{\pm 0.99}$ & $34.07$ & 3477 & 0.020 \\
        G-RAT & $8.45_{\pm 4.08}$ & $43.19$ & 6538 & 0.037 \\
        \method & $\sim$$36.00$ & $\sim$$180.00$ & 891 (\pred) & 0.005 \\ \midrule
        \multicolumn{5}{c}{HateXplain} \\ \midrule
        FR & $2.52_{\pm 0.30}$ & $13.49$ & 4324 & 122.0 \\
        MGR & $3.34_{\pm 0.13}$ & $18.11$ & 17032 & 480.6 \\ 
        MCD & $2.84_{\pm 0.29}$ & $15.11$ & 8452 & 238.5 \\
        G-RAT & $4.80_{\pm 0.43}$ & $25.39$ & 16840 & 475.2 \\
        \method & $\sim$$78.00$ & $\sim$$390.00$ & 2098 (\pred) & 59.2 \\ \bottomrule
    \end{tabular}
    \caption{Training running time and model size. We report single seed run running time (Single) and total running time over five seed runs (Total). Running time is measured in minutes. Additionally, we report the total number of trainable parameters and memory size.}
    \label{tab:appendix:time-and-size}
\end{table*}

\end{document}